\documentclass[runningheads]{llncs}

\usepackage{eccv}
\usepackage{eccvabbrv}
\usepackage{microtype}
\usepackage{graphicx}
\usepackage{xcolor}
\usepackage{booktabs}
\usepackage{tabularx}
\usepackage{xspace}
\usepackage{amsmath}
\usepackage{amssymb}
\usepackage{xurl}
\usepackage{float}

\newcommand{\best}[1]{\textbf{#1}}

\newcolumntype{Y}{>{\raggedright\arraybackslash}X}

\usepackage{placeins}
\usepackage[breaklinks,colorlinks,allcolors=eccvblue]{hyperref}

\title{Multi-Agent Target-Existence Verification\\
       and Learned Mask Geometry Refinement:\\
       Winning Report of the MeViS-Text Track\\
       at the 8th LSVOS Challenge 2026}
\titlerunning{Existence Verification and Geometry Refinement for MeViS-Text}

\author{Jungyoon Lee\inst{1} \and
Gyuil Lim\inst{2} \and
Doeon Kim\inst{3} \and
Seong-heum Kim\inst{1,2,3}\thanks{Corresponding author}}
\authorrunning{J.~Lee et al.}
\institute{
Department of AI Convergence Security, Soongsil University, Republic of Korea \and
Department of AI Convergence, Soongsil University, Republic of Korea \and
Department of Intelligent Semiconductors, Soongsil University, Republic of Korea\\
\email{\{jungyoon, gyuilLim, ilsin205\}@soongsil.ac.kr},
\email{seongheum@ssu.ac.kr}}

\newenvironment{paperwidefigure}[1][t]
  {\begin{figure}[#1]}{\end{figure}}
\newenvironment{paperwidetable}[1][t]
  {\begin{table}[#1]}{\end{table}}

\begin{document}
\raggedbottom
\setlength{\textfloatsep}{8pt plus 2pt minus 2pt}
\setlength{\floatsep}{8pt plus 2pt minus 2pt}
\setlength{\intextsep}{8pt plus 2pt minus 2pt}
\captionsetup{font=small,skip=3pt}
\maketitle

\begin{abstract}
We present the first-place solution to the MeViS-Text track of the 8th Large-scale Video Object Segmentation (LSVOS) Challenge 2026: referring video object segmentation guided by written motion expressions, including deceptive no-target expressions that match no object in the video and must yield empty masks in every frame.
Our pipeline, SSUPER, resolves each expression into a visual concept, generates full-video candidate masklets with SAM~3.1, and selects target IDs.
At every reasoning stage, three heterogeneous multimodal large language models independently execute the same stage-specific prompt before a single synthesis pass commits one schema-validated verdict. Although this system rejects every no-target expression in validation, the leaderboard reveals that a substantial share of test no-target cases still slips through. The reason is that hard negatives name a plausible object and fail only under the complete temporal predicate, so when selection and existence are decided together, a category-plausible masklet anchors the verdict.
Hence, we decouple existence verification into an independent multi-agent audit of the full predicate (category, count, action, trajectory, event order, and semantic role) that distinguishes absence from temporary invisibility, discounts apparent motion caused by camera movement, and requires contradicting evidence rather than mere uncertainty for a no-target verdict. Without any new segmentation call, this audit recovers most of the residual no-target errors. A training-data-only StyleRefiner then aligns mask geometry with the annotation style of MeViSv2 while preserving every presence decision by construction, showing that once the semantics are fixed, part of the remaining error is stylistic rather than semantic.
The complete system reaches a Final score of 0.9081339614 on the official challenge leaderboard.
\keywords{Referring video object segmentation \and Motion expressions \and
No-target recognition \and Multimodal large language models (MLLMs) \and
Mask geometry refinement}
\vspace{5mm}
\end{abstract}

\begin{paperwidefigure}[t]
\centering
\includegraphics[width=\textwidth,trim=0 80bp 0 0,clip]{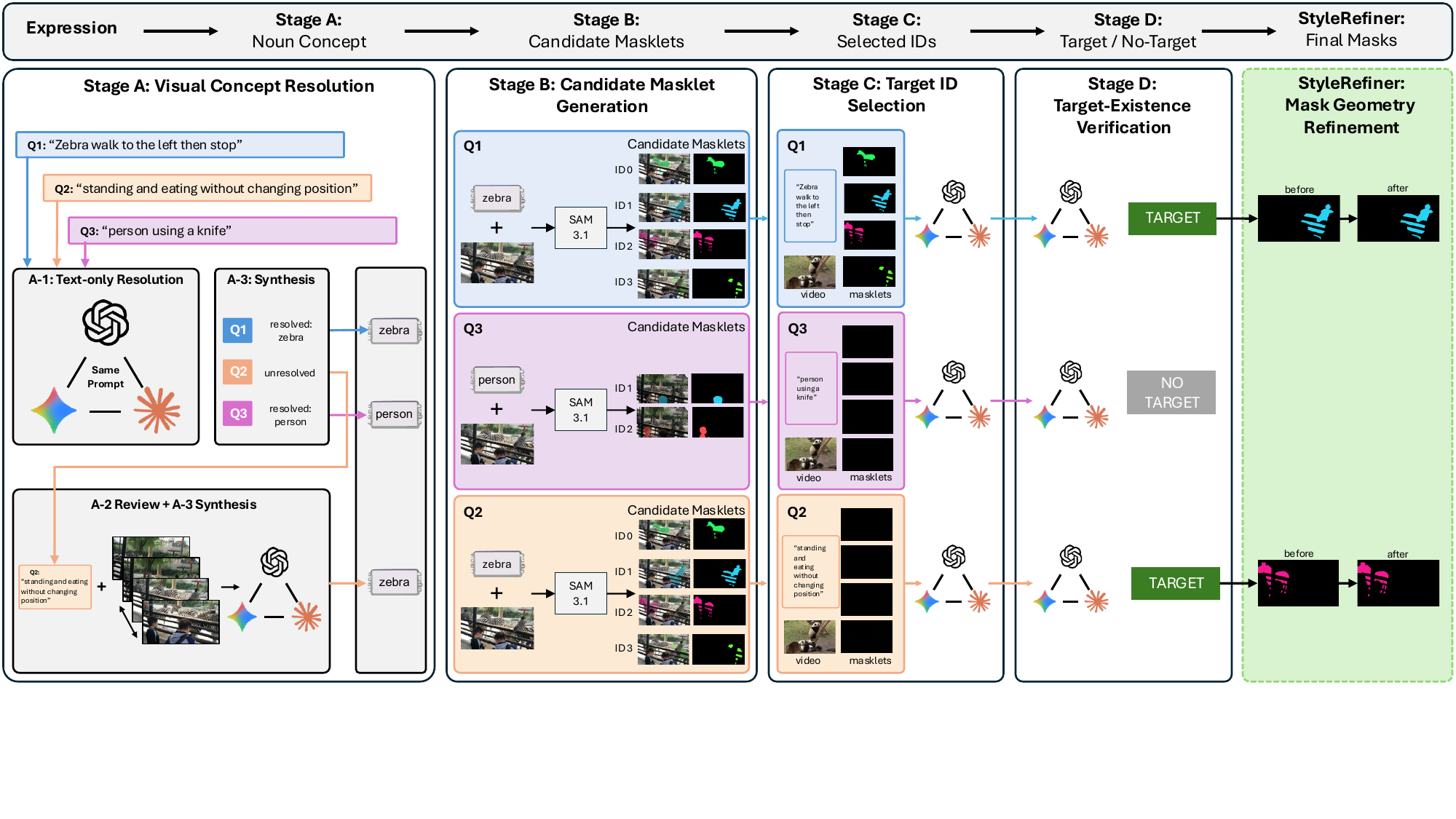}
\caption{\textbf{Overview of SSUPER, illustrated through three representative expressions.}
Q1 resolves directly to a noun concept from text alone; the noun-less Q2 requires ordered video evidence before a concept can be formed; and Q3 yields a category-plausible masklet that is later rejected because
the complete action predicate fails. At every reasoning stage, three heterogeneous MLLMs independently review the same stage-specific prompt and a synthesis pass commits a single verdict; if the text-only route of
Stage~A remains unresolved, the same block is repeated with visual evidence. Stage~D verifies target existence without any new segmentation call, and StyleRefiner refines only the geometry of masks that survive this existence verification.
}
\vspace{2mm}
\label{fig:v2-overview}
\end{paperwidefigure}

\section{Introduction}
\label{sec:v2-intro}
\vspace{-2mm}
The 8th Large-scale Video Object Segmentation (LSVOS) Challenge, held in conjunction with ECCV 2026, comprises three tracks: MOSEv2, MeViSv2-Text (MeViS-Text), and MeViSv2-Audio (MeViS-Audio). The first track evaluates VOS in complex scenes with severe occlusion and crowding~\cite{ding2025mosev2}, uses the MOSEv2 dataset~\cite{ding2025mosev2}. The latter two tracks use the text and audio modalities of MeViSv2~\cite{ding2025mevisv2}, respectively. Audio is speech-guided rather than sound-source segmentation. We participated only in the Text track. In the MeViS-Text track, on which we report, the target object or objects, if any, are specified by a written motion expression.

MeViS makes temporal evidence central to referring video object segmentation (RVOS)~\cite{ding2023mevis}: expressions describe how objects move and interact over time, so no single frame suffices to identify the referent. MeViSv2~\cite{ding2025mevisv2}, the official release of which we use~\cite{fudancvl2026mevisv2release}, sharpens this challenge with no-target expressions, deceptive motion descriptions that refer to no actual object in the video and must yield empty masks in every frame. Segmentation thereby becomes a joint problem of grounding and verification: a system must decide not only which track best matches the query, but whether any track truly satisfies it at all.

Recent systems couple multimodal large language models (MLLMs) with foundation segmenters, letting the language model reason over object tracks that the segmenter proposes~\cite{he2026strong,jin2026agentrvos}. We follow this track-before-selection design, but observe that a single reasoning model inherits its own blind spots at every stage. Our pipeline, SSUPER, instead runs three heterogeneous MLLMs in parallel on an identical stage-specific prompt at each reasoning stage, and commits a single schema-validated verdict through an explicit synthesis pass, so that no individual model's bias silently decides an expression. Figure~\ref{fig:v2-overview} overviews the pipeline.

Our validation results initially suggested that this system had solved no-target recognition, yet the challenge leaderboard disagreed: a substantial share of hidden-test no-target cases still slipped through. The
gap exposes a systematic failure mode rather than noise. Hard negatives name a plausible object and break only under the complete temporal predicate, so when candidate selection and target existence are decided in a single judgment, a category-plausible masklet anchors the verdict toward acceptance. The natural remedy is separation of concerns: we add a decoupled existence stage that audits the full predicate against the video, requires contradicting evidence rather than mere uncertainty for rejection, and touches no geometry, isolating existence reasoning from proposal generation. Its checklist is organized from labeled training expressions alone.

Our challenge report contributes:
\begin{itemize}
  \item A heterogeneous multi-agent workflow in which three MLLMs execute the same first-pass prompt for concept construction, track selection, and existence review, followed by an explicit synthesis pass.
  \item A decoupled target-existence verification stage, motivated by the observation that joint selection-and-existence judgments anchor on category-plausible candidates, which recovers most residual no-target
  errors without any new segmentation call.
  \item A train-only StyleRefiner that aligns mask geometry with the MeViSv2 annotation style while preserving every presence decision, showing that once the semantics are fixed, part of the remaining error is stylistic rather than semantic. The complete system reaches the track's top Final score.
\end{itemize}

\section{Method}
\label{sec:v2-method}

\paragraph{\textbf{Task.}}
Given a video $V=(I_1,\dots,I_T)$ and an expression $e$, we return masks $Y_e=(Y_{e,1},\dots,Y_{e,T})$. Singular and plural queries select one or all referred identities; a no-target query requires $Y_{e,t}=\varnothing$ for all $t$. Occlusion or late entry is frame-level absence, not no-target.

\subsection{Track-first grounding}
\label{sec:stagesac}

\paragraph{\textbf{Shared multi-agent protocol.}}
Stages A, C, and D use a review--synthesis block. OpenAI GPT-5.6~Sol, Anthropic Claude Fable~5, and Google Gemini 3.1~Pro receive identical stage-specific prompts/evidence and return schema-validated outputs. A
subsequent GPT-5.6~Sol call receives the evidence and the three responses in fixed order and commits one result under the same schema. It may resolve disagreement but cannot introduce a Stage~B candidate ID outside
the supplied inventory. Stages C and D use one block; Stage~A conditionally uses two.

\paragraph{\textbf{Visual evidence packaging.}}
Chronological contact sheets contain every metadata frame. Stage~A-2 starts with up to 14 uniformly spaced overview frames; Stages C and D use a 12-frame overview and numbered, color-coded 12-frame masklet overlays.
Within a stage, all agents receive identical ordered media whose tiles carry exact filenames, and must cite the frames used.

\paragraph{\textbf{Stage A: expression-wise visual concepts.}}
The three agents apply the same concept prompt to each expression. The call sequence is A-1 text-only review $\rightarrow$ A-3 GPT synthesis. If unresolved, A-2 repeats the review with ordered video evidence and A-3
synthesizes again. Thus direct and visually dependent cases use one and two blocks, respectively. A-3 returns a short noun phrase for SAM~3.1 or \texttt{unresolved}; action, direction, order, count, and relations remain
in the original expression for Stage~C. Figure~\ref{fig:v2-overview} illustrates both routes.

\paragraph{\textbf{Stage B: SAM 3.1 masklets.}}
The Stage~A concept prompts the SAM~3.1 video predictor with Object Multiplex~\cite{meta2026sam31}. Stage~B records returned full-video masklets with stable IDs, per-frame masks, boxes, and confidence, but does not choose the referent. Late appearance and occlusion remain represented by per-frame empty regions within a full-video candidate instead of forcing an expression-level rejection.

\paragraph{\textbf{Stage C: candidate-conditioned selection.}}
Stage~C receives the original expression, ordered RGB evidence, and
numbered masklets. The same selection prompt is executed independently by
the three agents, each of which compares appearance and temporal behavior
and returns one or more IDs, \texttt{no\_target}, or \texttt{unresolved}.
The synthesis pass commits the final decision. Its record stores both the
exported verdict and, when supported, the best non-empty candidate-ID set
before existence gating. Candidate selection and target existence are
therefore decided together, but the provisional IDs remain available to
Stage~D. The comparison uses the original motion expression, not only the
Stage~A noun.

\subsection{Diagnosing the existence gap}
\label{sec:diagnosis}

The Stage~C selection prompt explicitly permits a \texttt{no\_target} verdict; it is not an always-select baseline. It rejects all 38 no-target expressions of the \texttt{valid\_u} validation split, reaching an
N-accuracy (accuracy over no-target expressions) of 1.0000, whereas our first official challenge submission reaches only 0.8056 on the hidden test set. Given the small validation count, we treat this as a leaderboard-observed validation-to-challenge gap that motivated Stage~D, not as a pre-registered held-out ablation.

To understand the gap, we inspect training expressions along two axes. Surface form covers questions, passive voice, anaphora, and noun-less phrasing, and is handled mainly by the visual route of Stage~A. Predicate
structure covers count, state, trajectory, event order, and actor--patient binding, and can be tested only after tracking. Many hard negatives combine an obvious noun with a predicate that the video contradicts only in full, which is precisely the anchoring failure previewed in the introduction: once a category-plausible masklet exists, a joint selection-and-existence judgment tends to accept it.

\begin{paperwidetable}[t]
\caption{Train-derived predicate categories and their Stage D checks.}
\label{tab:v2-taxonomy}
\centering
\small
\setlength{\tabcolsep}{4pt}
\begin{tabularx}{\textwidth}{@{}lYY@{}}
\toprule
Predicate family & Typical Stage C failure & Stage D evidence \\
\midrule
Category/appearance & A similar instance is accepted & Verify literal category and attributes \\
Count/plural & One instance stands in for a set & Count qualifying identities over time \\
Action/state & Object lacks the stated behavior & Require the action or sustained state \\
Direction/trajectory & A short interval agrees, then reverses & Check full path; discount camera motion \\
Temporal composition & Only one sub-event occurs & Verify all clauses in order \\
Relation/semantic role & Actor, patient, or relation is reversed & Bind roles and test interaction \\
\bottomrule
\end{tabularx}
\end{paperwidetable}

\begin{paperwidefigure}[!t]
\centering
\vspace{5mm}
\includegraphics[width=\textwidth,trim=0 88bp 6bp 0,clip]{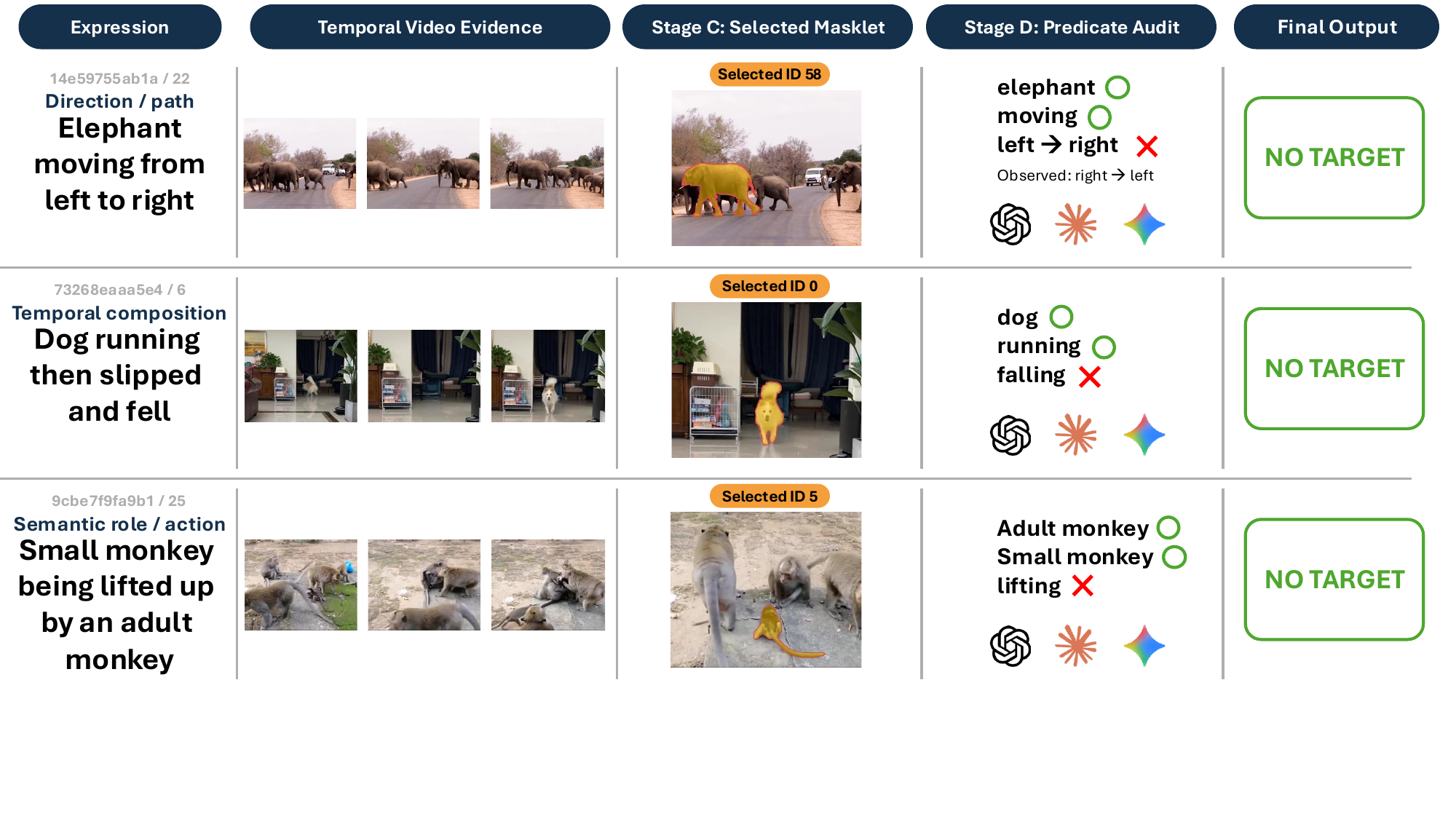}
\caption{\textbf{Three train-only no-target cases motivating predicate-aware
existence verification}. Top (trajectory): the elephant moves right to
left, contradicting the stated left-to-right path. Middle (temporal
composition): the dog runs, but the required slip-and-fall event never
occurs. Bottom (semantic role): the small monkey climbs onto the adult
rather than being lifted by it. Stage~C retains a category-plausible
masklet in every case, whereas Stage~D audits the complete predicate
against the full video and returns the correct empty output. Video and
expression IDs are shown at the top left.}
\vspace{3mm}
\label{fig:v2-audit}
\end{paperwidefigure}

Table~\ref{tab:v2-taxonomy} organizes the recurring predicate families into a multi-label taxonomy together with the corresponding existence checks. We build it from labeled training examples only; \texttt{valid\_u} serves for offline evaluation, never for constructing the taxonomy. Figure~\ref{fig:v2-audit} traces three train-only no-target cases for which a category-plausible masklet is insufficient, respectively exposing trajectory, temporal-composition, and actor--patient errors.

\subsection{Decoupled multi-agent existence verification}
\label{sec:staged}

Stage~D asks: \emph{does any object in the complete video satisfy the full expression?} The three agents independently receive the same existence-verification prompt and ordered evidence, then produce structured target/no-target verdicts with cited frames and predicate rationales. Their outputs are finalized by the shared synthesis pass. The initial reviews do not receive the Stage~C verdict.

Reviews cover the beginning, middle, and end of the video and distinguish absence from temporary invisibility. A no-target decision requires evidence that contradicts one of the predicate families of Table~\ref{tab:v2-taxonomy}: a required category, appearance, count, action, trajectory, event order, or semantic role. Screen displacement alone does not establish object motion under camera pan or zoom. Agents cite checked frames and predicates. Uncertainty is not converted to no-target; the synthesis pass resolves conflicts or retains \texttt{unresolved} when the evidence is insufficient.

Stage~D generates no geometry. A \texttt{no\_target} verdict exports an empty sequence. A target verdict preserves a non-empty Stage~C result. If Stage~C exported an empty sequence, Stage~D may restore only the same expression's archived provisional candidate IDs; it never borrows a selection from another expression. Without such a non-empty record, the case remains \texttt{unresolved} and empty. Stage~D never calls SAM or creates a candidate ID. Thus the A--C to A--D comparison isolates existence reasoning from new proposal generation.

\subsection{StyleRefiner: mask geometry after target selection}
\label{sec:refiner}

StyleRefiner learns to align SAM-derived mask geometry with the MeViSv2 annotation geometry after all semantic decisions are fixed. For each visible object-frame in the MeViSv2 training split, three annotation-derived interior points and a tight box prompt a frozen SAM~3.1 image predictor. We select the points deterministically by repeatedly taking a distance-transform maximum and suppressing its surrounding disk. The SAM mask prompted from the corresponding annotation is paired with that annotation: normalized RGB and the SAM mask form the
model input, and the annotation is the training target. Filtering objects below 64 pixels and empty SAM outputs yields 369,696 pairs. We split videos, rather than frames, into 1,413 training and 249 development videos so that adjacent frames cannot cross the split. Neither \texttt{valid\_u} nor test annotations are used for training or checkpoint selection.

We train separate models at $R\in\{512,768\}$; they do not share a checkpoint. The predicted-mask box is expanded by 25\%, made square, and resized to $R\times R$. The five-channel input is normalized RGB, the mask prior, and its signed distance transform. We expand the pretrained ConvNeXt-S stem from three to five channels: RGB slices inherit its initialization and the two added slices are zero-initialized. A high-resolution detail stem is learned from scratch on all five channels. The DINOv3-distilled encoder~\cite{liu2022convnext,simeoni2025dinov3}, detail stem, and U-Net-like decoder~\cite{ronneberger2015unet} contain 56.6M parameters. We train with boundary-weighted BCE, soft Dice, differentiable boundary-F, and auxiliary losses.

AdamW uses encoder/decoder learning rates $10^{-4}/4{\times}10^{-4}$, encoder layer decay 0.8, weight decay 0.05, 300-step warm-up, cosine decay, gradient clipping 1.0, and EMA 0.999. Weighted BCE, Dice, and boundary-F have weights 1, 1, and 0.5; stride-4/8 auxiliary BCE--Dice losses have weights 0.4/0.2. On eight GPUs, the 512/768 runs use per-GPU batches 32/16; their scored EMA checkpoints are selected after completed epochs three and two on the same 400 development pairs. The 768 run completes three epochs. Augmentation uses 0.5-probability horizontal
flips, scale 0.85--1.30, $\pm15^{\circ}$ rotation, $\pm8\%$ translation, and photometric jitter. With probability 0.5, one or two synthetic prior corruptions are additionally chosen from erosion/dilation (radius 1--3), translation (up to 4 pixels), elastic warping, and boundary-blob insertion or removal.

At inference, the prior is the assembled Stage~D mask rather than a ground-truth-prompted image mask, and each checkpoint uses its matching crop resolution. Each connected component of at least 64 pixels iscropped, refined, and pasted at original resolution before components are reunited. Empty frames and smaller components are copied, and an erased foreground falls back to its non-empty input. No ground truth, text, candidate ID, or new SAM call is used. A final invariant enforces the Stage~D expression-level target-presence decision.

\section{Experiments}
\label{sec:experiments}

\paragraph{\textbf{Data.}}
We count expressions directly from the official Hugging Face MeViSv2
release used in our experiments~\cite{fudancvl2026mevisv2release}. Its training metadata contains 1,662 videos and 27,502 expressions: 3,778 no-target and 23,724
target. The release also provides \texttt{valid\_u}: 50 videos and 907 expressions, comprising 38 no-target and 869 target cases. The challenge test set has 50 videos and 444 expressions; its annotations are hidden, and we report only official aggregate metrics. The release page lists 2,006 videos and 33,458 sentences in total; the splits above account for 1,762 videos, and we do not use the remaining official split.
\paragraph{\textbf{Metrics.}}
The challenge reports region similarity $\mathcal{J}$, contour accuracy $\mathcal{F}$, their mean $\mathcal{J}\&\mathcal{F}$, N-accuracy over no-target expressions, and T-accuracy over target expressions. The ranking metric is

\begin{equation}
  \mathrm{Final} \;=\; \tfrac{1}{3}\bigl(\mathcal{J}\&\mathcal{F}
  + \text{N-acc} + \text{T-acc}\bigr).
  \label{eq:final}
\end{equation}

\paragraph{\textbf{Configuration.}}
Stages A, C, and D use OpenAI GPT-5.6 Sol (\texttt{gpt-5.6-sol}, reasoning effort \texttt{max}), Anthropic Claude Fable 5 (\texttt{claude-fable-5}, effort \texttt{high}), and Google Gemini 3.1 Pro (\texttt{gemini-3.1-pro-preview}, thinking level \texttt{high}). All agents in an independent review receive the same prompt and evidence before GPT-5.6~Sol synthesis; Stage~A repeats the block only if visually unresolved. Stage~B uses the SAM~3.1 video predictor with Object
Multiplex. The 512- and 768-pixel StyleRefiners are trained separately at their respective resolutions; the final system uses the 768-pixel EMA checkpoint after completed epoch two (zero-indexed epoch~1), selected on the video-disjoint train--development split. Test values are official CodaBench aggregates. Stage~D receives the unchanged Stage~A--C output and makes no new SAM call.

\subsection{Official stage-wise results}
\label{sec:results}

\begin{paperwidetable}[t]
\caption{Official stage-wise aggregate results. The 512- and 768-pixel StyleRefiners are independently trained resolution-specific models.}
\label{tab:v2-primary}
\centering
\small
\setlength{\tabcolsep}{4pt}
\begin{tabular}{@{}llrrrr@{}}
\toprule
Split & System & $\mathcal J\&\mathcal F$ & N-acc. & T-acc. & Final \\
\midrule
\texttt{valid\_u} & Stage A--C & .841901 & 1.000000 & .985040 & .942314 \\
Test & Stage A--C & .760935 & .805556 & .977941 & .848144 \\
Test & Stage A--D & .778701 & .944444 & .987745 & .903630 \\
Test & + Refiner 512 & .790788 & .944444 & .987745 & .907659 \\
Test & + Refiner 768 & \best{.792212} & \best{.944444} & \best{.987745} & \best{.908134} \\
\bottomrule
\vspace{2mm}
\end{tabular}
\end{paperwidetable}

Table~\ref{tab:v2-primary} gives official aggregates. Relative to Stage~A--C, Stage~D improves N-accuracy by 0.1389, T-accuracy by 0.0098, and Final by 0.0555. Its N-accuracy corresponds to 34 correct no-target decisions among the 36 test no-target expressions. Stage~D creates no geometry, but suppressing an export or restoring same-expression provisional IDs changes which existing masklet sequence is scored, explaining the 0.0178 $\mathcal{J}\&\mathcal{F}$ gain; the T-accuracy
gain likewise stems from restored provisional IDs, since suppression alone could only lower it.

\subsection{Mask geometry refinement}
\label{sec:geom}

StyleRefiner preserves every N/T decision. At 512 pixels it reaches
$\mathcal{J}=.764889$ and $\mathcal{F}=.816686$. At 768 pixels, it raises
$\mathcal{J}$ from .758950 to .765711 and $\mathcal{F}$ from .798451 to
.818713 relative to Stage~D. This adds 0.0135 $\mathcal{J}\&\mathcal{F}$
and 0.0045 Final, with the larger $\mathcal{F}$ gain supporting its
contour-focused role.

\begin{paperwidefigure}[t]
\centering
\includegraphics[width=\textwidth,trim=0 34bp 130bp 4bp,clip]{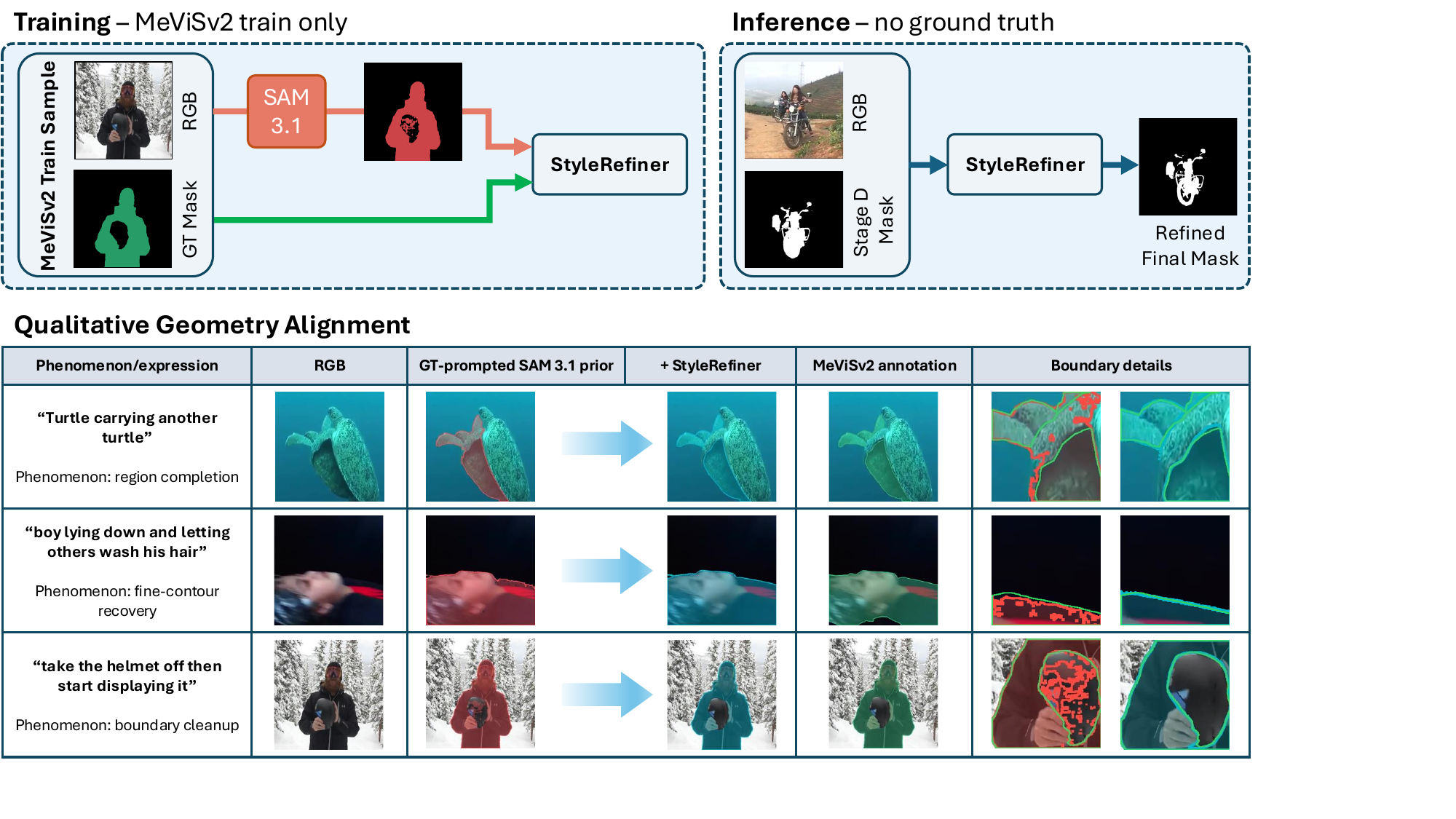}
\caption{\textbf{StyleRefiner training, inference, and qualitative geometry
alignment.} Top left (training): for each MeViSv2 training frame, the
ground-truth annotation prompts SAM~3.1, and the resulting mask is paired
with that annotation to supervise annotation-aligned refinement. Top right
(inference): the trained model refines assembled Stage~D masks without any
ground truth, preserving the expression-level empty/non-empty decision.
Bottom: three held-out development examples from the video-disjoint
training split compare the SAM prior, the refined mask, and the MeViSv2
annotation, illustrating region completion, fine-contour recovery, and
boundary cleanup, respectively.}
\label{fig:v2-style}
\vspace{2mm}
\end{paperwidefigure}

For qualitative evaluation, Figure~\ref{fig:v2-style} contrasts StyleRefiner
training with deployment and shows three held-out examples from the
video-disjoint MeViSv2 training split. Their priors are
ground-truth-prompted masks used to construct training pairs rather than
deployed Stage~D masks. The side-by-side prior, refined mask, annotation, and
boundary crops isolate the learned geometry correction on unseen videos.

Each row isolates one recurring failure mode of raw SAM geometry. In the
region-completion case, one turtle carries another of the same category,
and the prior truncates the referred instance where the two bodies
overlap; the refiner restores the annotated extent. In the fine-contour-recovery case, the hair being washed forms a thin, low-contrast structure that the prior over-smooths, and the refiner re-carves the contour at the level of detail the annotations maintain. In the boundary-cleanup case, the prior follows the target with a ragged outline that the refiner snaps to the tight annotation-style boundary.

Three observations follow. First, the corrections are contour-dominated:
regions change little while boundaries move substantially, matching the
larger $\mathcal{F}$ gain. Second, each failure mode has a training-time counterpart among the synthetic prior corruptions: erosion and blob removal emulate missing regions, elastic warping emulates contour drift, and boundary-blob edits emulate ragged outlines. This correspondence is what lets the learned correction carry over to deployed Stage~D priors, although the gap between the two prior types remains. Third, in every example the refinement changes how the target is drawn, never whether it exists; the residual error it removes is stylistic, not semantic.

\subsection{Limitations}
\label{sec:limitations}
In our current implementation, Stage~D cannot recover geometry absent from Stage~B. Hosted-model drift means replay requires archived prompts, responses, and model metadata, while multi-model review adds cost. Aggregate test metrics preclude
predicate-stratified analysis. Only StyleRefiner is trained on MeViSv2; it improves the mean but retains a regression tail. Its annotation-prompted train priors differ from deployed Stage~D masklets. Synthetic corruption mitigates this gap, but does not guarantee improvement for every mask.

\section{Conclusion}
\label{sec:conclusion}
Our winning MeViS-Text solution, SSUPER, uses heterogeneous MLLMs to construct concepts, select masklets, and verify target existence. Stage~D reaches a test N-accuracy of 0.9444 without a new SAM run; StyleRefiner preserves presence decisions while raising $\mathcal{J}\&\mathcal{F}$ to 0.792212. The Final score (the official ranking metric, averaging $\mathcal{J}\&\mathcal{F}$, N-accuracy, and T-accuracy) is 0.9081339614 on the challenge leaderboard.

\section*{Acknowledgements}
This research was supported by G-LAMP Program of the National Research Foundation of Korea (NRF) grant funded by the Ministry of Education (No.~RS-2025-25441317). This work was also supported by the NRF grant funded by the Ministry of Science and ICT (MSIT) (RS-2025-16071992), and by MSIT and the National IT Industry Promotion Agency through the Advanced GPU Utilization Support Program (02-26-01-0499). This work was also supported by Korea Institute for Advancement of Technology(KIAT) grant funded by the Korea Government(MOTIE) (RS-2026-25530975, HRD Program for Industrial Innovation), and the MSIT(Ministry of Science and ICT), Korea, under the Convergence security core talent training business support program(IITP-2024 2024-RS-2024-00426853) supervised by the IITP(Institute of Information \& Communications Technology Planning \& Evaluation).

\bibliographystyle{splncs04}
\bibliography{ssuper}

@inproceedings{ding2023mevis,
  author    = {Henghui Ding and Chang Liu and Shuting He and Xudong Jiang and Chen Change Loy},
  title     = {{MeViS}: A Large-Scale Benchmark for Video Segmentation with Motion Expressions},
  booktitle = {Proceedings of the IEEE/CVF International Conference on Computer Vision},
  pages     = {2694--2703},
  year      = {2023}
}

@article{ding2025mevisv2,
  author  = {Henghui Ding and Chang Liu and Shuting He and Kaining Ying and Xudong Jiang and Chen Change Loy and Yu-Gang Jiang},
  title   = {{MeViS}: A Multi-Modal Dataset for Referring Motion Expression Video Segmentation},
  journal = {IEEE Transactions on Pattern Analysis and Machine Intelligence},
  volume  = {47},
  number  = {12},
  pages   = {11400--11416},
  year    = {2025},
  doi     = {10.1109/TPAMI.2025.3600507}
}

@misc{fudancvl2026mevisv2release,
  author       = {{FudanCVL}},
  title        = {{MeViSv2}: Official Dataset Release},
  howpublished = {Hugging Face Datasets, \url{https://huggingface.co/datasets/FudanCVL/MeViSv2}},
  note         = {Release metadata accessed July 2026},
  year         = {2026}
}

@article{ding2025mosev2,
  author  = {Henghui Ding and Kaining Ying and Chang Liu and Shuting He and Xudong Jiang and Yu-Gang Jiang and Philip H. S. Torr and Song Bai},
  title   = {{MOSEv2}: A More Challenging Dataset for Video Object Segmentation in Complex Scenes},
  journal = {arXiv preprint arXiv:2508.05630},
  year    = {2025}
}

@article{jin2026agentrvos,
  author  = {Woojeong Jin and Jaeho Lee and Heeseong Shin and Seungho Jang and Junhwan Heo and Seungryong Kim},
  title   = {{AgentRVOS}: Reasoning over Object Tracks for Zero-Shot Referring Video Object Segmentation},
  journal = {arXiv preprint arXiv:2603.23489},
  year    = {2026}
}

@article{he2026strong,
  author  = {Xusheng He and Canyang Wu and Jinrong Zhang and Weili Guan and Jianlong Wu and Liqiang Nie},
  title   = {The 1st Winner for 5th {PVUW} {MeViS-Text} Challenge: Strong {MLLMs} Meet {SAM3} for Referring Video Object Segmentation},
  journal = {arXiv preprint arXiv:2604.00404},
  year    = {2026}
}

@misc{meta2026sam31,
  author       = {{Meta AI}},
  title        = {{SAM 3.1}: Faster and More Accessible Real-Time Video Detection and Tracking with Multiplexing and Global Reasoning},
  howpublished = {\url{https://ai.meta.com/blog/segment-anything-model-3/}},
  note         = {Accessed July 26, 2026},
  year         = {2026}
}

@inproceedings{liu2022convnext,
  author    = {Zhuang Liu and Hanzi Mao and Chao-Yuan Wu and Christoph Feichtenhofer and Trevor Darrell and Saining Xie},
  title     = {A ConvNet for the 2020s},
  booktitle = {Proceedings of the IEEE/CVF Conference on Computer Vision and Pattern Recognition},
  pages     = {11976--11986},
  year      = {2022}
}

@article{simeoni2025dinov3,
  author  = {Oriane Sim{\'e}oni and Huy V. Vo and Maximilian Seitzer and Federico Baldassarre and Maxime Oquab and Cijo Jose and Vasil Khalidov and Marc Szafraniec and others},
  title   = {{DINOv3}},
  journal = {arXiv preprint arXiv:2508.10104},
  year    = {2025}
}

@inproceedings{ronneberger2015unet,
  author    = {Olaf Ronneberger and Philipp Fischer and Thomas Brox},
  title     = {{U-Net}: Convolutional Networks for Biomedical Image Segmentation},
  booktitle = {Medical Image Computing and Computer-Assisted Intervention},
  pages     = {234--241},
  year      = {2015}
}

\end{document}